\documentclass[11pt]{article}

\usepackage[preprint]{acl}
\usepackage{times}
\usepackage{latexsym}
\usepackage[T1]{fontenc}
\usepackage[utf8]{inputenc}
\usepackage{microtype}
\usepackage{inconsolata}
\usepackage{graphicx}
\usepackage{xcolor} % 颜色支持
\usepackage{booktabs}
\usepackage{multirow}
\usepackage{amsmath}
\usepackage{algorithm}
\usepackage{algorithmicx}
\usepackage{algpseudocode}
\usepackage{enumitem}
\usepackage[most]{tcolorbox}
\usepackage{listings}

\newcommand{\robusthl}[1]{\colorbox{blue!12}{\strut #1}}
\newcommand{\metaahl}[1]{\colorbox{red!12}{\strut #1}}

\title{Evaluating Interactive Reasoning in Large Language Models: A Hierarchical Benchmark with Executable Games}

\author{
 \textbf{Mingyuan Fan\textsuperscript{1}},
 \textbf{Weiguang Han\textsuperscript{2}},
 \textbf{Daixin Wang\textsuperscript{2}},
 \textbf{Cen Chen\textsuperscript{1}},
\\
 \textbf{Zhiqiang Zhang\textsuperscript{2}},
 \textbf{Jun Zhou\textsuperscript{2}}
\\
\\
 \textsuperscript{1}East China Normal University,
 \textsuperscript{2}Ant Group
}

\begin{document}
\maketitle

\begin{abstract}
We introduce a multi-turn interactive framework for reasoning evaluation that treats reasoning as active evidence acquisition and belief updating.
Wherein, LLMs receive only the task rules, must issue targeted queries to a hidden environment, integrate partial observations over time, and decide when to submit a final answer.
Beyond standard success rate and interaction efficiency, we evaluate \emph{contextual robustness} under controlled contextual perturbations, and \emph{metacognitive adaptation} through \emph{counterfactual revision} and \emph{necessity judgment}.
We instantiate the framework as a benchmark of 474 executable games, each evaluated under five fixed configuration search spaces corresponding to five difficulty levels, and evaluate a broad set of frontier LLMs.
Results show that the benchmark is highly discriminative, exposing large differences not only in success rate but also in interaction efficiency.
Moreover, we empirically show that contextual perturbations cause moderate but consistent declines, whereas counterfactual revision and necessity judgment lead to much larger drops. 
% Overall, current LLMs show meaningful interactive reasoning ability but remain limited in robustness, belief revision, and evidence attribution.
\end{abstract}

\section{Introduction}

Large language models (LLMs) have recently achieved impressive results on reasoning benchmarks \cite{gemini,deepseekv3}. Yet most existing evaluations remain fundamentally static \cite{ref9,ref5,ref26}: the model is given a fully specified problem and asked to produce a final answer in a single shot. This setup is increasingly insufficient for evaluating reasoning itself. On the one hand, it does not test whether a model can actively seek missing information, update beliefs over time, and decide when evidence is sufficient. On the other hand, failures on static benchmarks often conflate two distinct sources \cite{bang2025hallulens,yin2025reasoning}: a knowledge deficit, where the model lacks the necessary facts, and a reasoning deficit, where it has enough information but fails to use it correctly. Because static evaluations rarely separate these cases, their diagnostic value is limited.

To address this limitation, we propose a hierarchical framework for interactive reasoning evaluation that treats the model as an active agent under partial observability. Instead of receiving the full problem specification upfront, the model is given only the task rules and must iteratively issue targeted queries, gather partial evidence, update its beliefs, and decide when to submit an answer. This setting more directly measures reasoning as sequential information acquisition and evidence integration, while also reducing contamination concerns because there is no single fixed prompt-answer pair to memorize.

To isolate reasoning from factual recall and semantic priors, we build the games from minimal structural primitives. Specifically, we define hidden state spaces over four canonical data structures (sets, sequences, trees, and graphs) and instantiate them across three inference modes: deductive, inductive, and abductive. This structural design yields a controlled testbed in which performance can be attributed more cleanly to algorithmic reasoning rather than real-world knowledge.

On top of this interactive backbone, we introduce two higher-order evaluation layers to obtain a more fine-grained view of model capability. The first, \emph{contextual robustness}, tests whether reasoning is preserved under semantic perturbation, irrelevant context, and shifts in episode boundaries. The second, \emph{metacognitive adaptation}, examines whether models can revise beliefs when earlier evidence is corrected and whether they can distinguish logically necessary information from merely sufficient information. %Together, these probes move evaluation beyond final-answer accuracy toward a more diagnostic account of how reasoning succeeds or fails.

We instantiate this framework as a benchmark of 474 executable games, each evaluated under five difficulty levels (i.e., five different configuration search spaces), for a total of 2370 instances. The benchmark covers all combinations of the four data structures and three inference modes, and includes controlled probes for contextual robustness and metacognitive adaptation. We further note that several recent benchmarks~\cite{ref41,Multi-Turn-Puzzles,GTBENCH,GameArena} have explored multi-turn reasoning in interactive settings such as games or dialogue. However, these benchmarks are often constrained in terms of scale, structural diversity, or the extent to which they can disentangle distinct components of reasoning ability.

Empirically, we find that the benchmark is highly discriminative: frontier models differ substantially not only in final success rate but also in interaction efficiency. Performance varies systematically across reasoning regimes, with deductive tasks generally easier than abductive ones and set-based games emerging as the most challenging family. We also find that contextual perturbations cause moderate degradation, whereas counterfactual revision and evidence-pruning probes reveal much larger weaknesses. These results suggest that current LLMs are better at solving problems under fixed evidence than at maintaining a revisable account of why an answer is correct.
Our main contributions are as follows:
\begin{itemize}[leftmargin=*]
    \item We propose a hierarchical framework for evaluating interactive reasoning, casting reasoning as active information acquisition under partial observability rather than one-shot answer generation. Moreover, we introduce two higher-order evaluation layers, contextual robustness and metacognitive adaptation, to probe abstraction invariance, noise tolerance, boundary control, belief revision, and necessity judgment.
    \item We construct a benchmark of 474 executable games spanning four canonical data structures and three inference modes, with five fixed configuration search spaces corresponding to five difficulty levels, enabling controlled, comparable, and contamination-resistant evaluation.
    \item We provide a systematic empirical study of frontier LLMs, showing that while current models can often perform interactive search, they remain substantially weaker in robustness, belief revision, and evidence attribution.
\end{itemize}

\section{Related Work}

A large body of prior work evaluates LLM reasoning in static, single-turn settings, where the model receives a fully specified problem and produces a final answer. Logical reasoning benchmarks such as ReClor and LogiQA expose the gap between shallow pattern matching and deductive inference \cite{ref9,ref10}. Mathematical benchmarks such as GSM8K, MATH, MathVista, FrontierMath, and Humanity’s Last Exam test increasingly difficult forms of problem solving \cite{ref5,ref18,ref20,ref22,ref23}. Code reasoning has similarly progressed from function synthesis benchmarks such as HumanEval and APPS \cite{ref26,ref27} to more realistic software-engineering tasks such as SWE-bench \cite{ref30}.

Recent benchmarks move toward multi-turn evaluation. MT-Bench and follow-up datasets such as MT-Bench-101, MultiChallenge, and TurnBench-MS assess abilities including long-horizon instruction following and multi-step reasoning \cite{ref36,ref38,ref40,ref41,ref42}. However, these settings are mostly conversation-centric: models respond to evolving prompts rather than actively querying a hidden environment for evidence. They also often mix reasoning with world knowledge, dialogue ability, and instruction following.

Closest to our work are game-based and interactive benchmarks. MTR-Bench \cite{ref41} evaluates multi-turn interaction across 40 automated tasks. Multi-Turn Puzzles \cite{Multi-Turn-Puzzles} studies dialogue-based puzzle solving and logical consistency. GTBench \cite{GTBENCH} evaluates strategic reasoning in board and card games, while GameArena \cite{GameArena} uses live human-LLM interaction in Roblox mini-games to probe deductive and inductive reasoning.

Overall, these work has begun to study multi-turn reasoning, but rarely isolates reasoning as sequential evidence acquisition and belief updating. It also seldom tests robustness to controlled contextual perturbations or adaptation after earlier evidence is revised. Many benchmarks further conflate reasoning with factual knowledge, dialogue skill, or task-specific strategy, or remain limited in scale. Our work is designed to address these limitations.

\section{A Hierarchical Evaluation Framework for Interactive Reasoning} 

We present a hierarchical framework for interactive reasoning evaluation. At its core is a clean interactive backbone that formulates reasoning as state search under partial observability. On top of this backbone, we introduce two higher-level evaluation layers, namely contextual robustness and metacognitive adaptation, to probe increasingly advanced capabilities. Sections \ref{sec:design_principle}$\sim$\ref{sec:metacognitive_adaptation} focus on the conceptual design of these layers; the executable benchmark instantiation and construction pipeline are described in Section \ref{sec:instantiation}.

\subsection{Design Principles}
\label{sec:design_principle}

Our framework is guided by four principles.
\begin{itemize}[leftmargin=*]
    \item \textbf{Resistance to contamination.} Static benchmarks expose fixed prompt-answer pairs and are therefore vulnerable to memorization. By contrast, our evaluation is interactive and dynamically instantiated: each episode is generated from a hidden configuration, and the evidence revealed depends on the model's own actions. This greatly reduces the chance of memorizing fixed trajectories or answer patterns.
    \item \textbf{Hierarchical diagnostic design.} We begin with basic interactive search under uncertainty, then add progressively harder conditions. This layered design helps distinguish failures of basic search, contextual robustness, and adaptive revision.
    \item \textbf{Discriminative evaluation.} A useful benchmark should separate models of different capability levels. We vary difficulty through five fixed configuration search spaces, which control search-space size and hidden-state complexity while preserving comparability across models, methods, and evaluation settings.
    \item \textbf{Structural diversity and coverage.} Rather than relying on a single task template, we build games across multiple data structures and inference types to test whether reasoning generalizes across qualitatively different interactive settings.
\end{itemize}

\begin{algorithm}[t]
\caption{Interactive Protocol}
\label{alg:interaction_protocol}
\footnotesize
\textbf{Input:} Game Type $\mathcal{Q}$, Game Configuration $\mathcal{C}$, LLM $\pi$, Max turn budget $T_{\max}$ \\
\textbf{Output:} Final Status $E_{\text{status}} \in \{\text{Success, Failure, FormatError, Timeout}\}$, Interaction Count $N$
\begin{algorithmic}[1]
\State $\mathcal{E} \gets \text{InstantiateGame}(\mathcal{Q}, \mathcal{C})$
\State $p_0 \gets \mathcal{E}.\text{getRules}()$
\State \robusthl{$p_0 \gets \text{ContextWrapper}(p_0)$} \Comment{Optional: isomorphic perturbation / rule noise}
\State $H_0 \gets [p_0]$ \Comment{Initialize interaction history}
\State \robusthl{$H_0 \gets \text{HistoryWrapper}(H_0)$} \Comment{Optional: inter-game boundary control}
\For{$t = 1, 2, \dots, T_{\max}$}
    \State $r_t^{\text{agent}} \gets \pi(H_{t-1})$ \Comment{LLM acts on full history}
    
    \If{$\mathcal{E}.\text{isQuery}(r_t^{\text{agent}})$}
        \If{$\mathcal{E}.\text{is\_invalid\_format}(r_t^{\text{agent}})$}
            \State \Return (\text{FormatError}, $t$)
        \EndIf
        \State $r_t^{\text{env}} \gets \mathcal{E}.\text{respondToQuery}(r_t^{\text{agent}})$
        \State \robusthl{$r_t^{\text{env}} \gets \text{NoiseWrapper}(r_t^{\text{env}})$} \Comment{Optional: intra-game noise injection}
        \State \metaahl{$(r_t^{\text{env}}, H_{t-1}) \gets \text{RevisionWrapper}(r_t^{\text{env}}, H_{t-1})$} \Comment{Optional: counterfactual revision}
        \State $H_t \gets H_{t-1} \oplus [r_t^{\text{agent}}, r_t^{\text{env}}]$
    
    \ElsIf{$\mathcal{E}.\text{isSubmit}(r_t^{\text{agent}})$}
        \If{$\mathcal{E}.\text{is\_invalid\_format}(r_t^{\text{agent}})$}
            \State \Return (\text{FormatError}, $t$)
        \EndIf
        \If{$\mathcal{E}.\text{checkAnswer}(r_t^{\text{agent}})$}
            % \State \metaahl{$\text{OptionalNecessityProbe}(H_t)$} \Comment{sufficiency vs.\ necessity}
            \State \Return (\text{Success}, $t$)
        \Else
            \State \Return (\text{Failure}, $t$)
        \EndIf
    \Else
        \State \Return (\text{FormatError}, $t$)
    \EndIf
\EndFor
\State \Return (\text{Timeout}, $T_{\max}$)
\end{algorithmic}
\end{algorithm}

\subsection{Interactive Reasoning Backbone}
\label{sec:backbone}

At the core of our framework, each task is formulated as a multi-turn interactive game under partial observability. An LLM interacts with an environment containing a hidden state and must infer the target answer by actively gathering information. Unlike standard reasoning benchmarks, where all information is given upfront, our setting initially provides only the game rules; the model must reduce uncertainty through querying and evidence integration.

As shown in Algorithm~\ref{alg:interaction_protocol}, an episode begins by instantiating an environment $\mathcal{E}$ from game type $\mathcal{Q}$ and configuration $\mathcal{C}$. This separation lets us vary difficulty while preserving task structure. The environment returns an initial prompt $p_0=\mathcal{E}.\mathrm{getRules}()$, which specifies the objective and the valid action format, and the interaction history is initialized as $H_0=[p_0]$. At each step $t$, the model generates a response $r_t^{\text{agent}}=\pi(H_{t-1})$. The environment parses this response as either a query or a final submission. Queries are answered by $\mathcal{E}.\mathrm{respondToQuery}(\cdot)$ and appended to the history; submissions are evaluated by $\mathcal{E}.\mathrm{checkAnswer}(\cdot)$, yielding \textsc{Success} or \textsc{Failure}. Unparseable responses are labeled \textsc{FormatError}. If no valid submission is made within the budget $T_{\max}$, the episode ends with \textsc{Timeout}.

Note that our parsing approach follows a structured-action design to simplify parsing and ensure consistent evaluation, as it allows the model's responses to be processed reliably and efficiently. For example, a submission takes the form \texttt{<submit>answer</submit>}, which allows the environment to extract the proposed answer through exact pattern matching rather than brittle free-form interpretation. We report two main metrics: \emph{Success Rate}, i.e., the fraction of episodes that end in \textsc{Success}, and \emph{Avg.\ Turns} computed over successful episodes.

\textbf{Remark.}
The backbone defines a clean reference setting: rules are explicit, history contains only task-relevant content, and environment responses are faithful and noise-free. The two higher-order layers are systematic departures from this baseline along two axes. Contextual robustness preserves the latent search problem while perturbing surface form or surrounding context. Metacognitive adaptation changes the informational dynamics of the protocol itself, for example by revising earlier evidence or asking which observations are actually necessary for the conclusion.

\subsection{Contextual Robustness}
\label{sec:contextual_abstraction}

Real-world reasoning rarely appears in such purified form. To this end, we introduce a first higher-order evaluation layer centered on contextual robustness in the following probes:
\begin{itemize}[leftmargin=*]
    \item \textbf{Isomorphic perturbation}. Starting from the clean symbolic games, we construct semantically enriched variants that preserve the underlying data structure while changing only the surface interpretation of entities and relations. For example, graph nodes may become patients in a contact network. As these variants are structurally isomorphic to the original game, the search problem and the correct solution remain unchanged.
    \item \textbf{Intra-game noise injection}. We augment the rules and environment responses with non-informative text or irrelevant background descriptions while preserving the evidential content of the interaction. This probe tests whether redundant context degrades evidence tracking and weakens reasoning consistency.
    \item \textbf{Inter-game boundary control}. After the completion of one game, a second game is introduced within the same ongoing context without resetting the interaction history. The model must recognize the episode boundary, discard no-longer-relevant prior context, and reason about the new game from a clean latent state.
\end{itemize}
We measure contextual robustness by comparing performance under each perturbation against the corresponding clean condition.

\subsection{Metacognitive Adaptation}
\label{sec:metacognitive_adaptation}

Beyond effective search and contextual invariance, we ask whether a model can revise its reasoning when accepted information changes, and whether it can distinguish sufficient from necessary evidence. These abilities probe whether the model maintains an internal account of why its conclusion is justified.
\begin{itemize}[leftmargin=*]
    \item \textbf{Counterfactual revision}. During an ongoing game, after the model has already accumulated a nontrivial body of evidence, the environment issues a correction to one earlier observation. The model must then update its beliefs and continue the search accordingly. A strong model should revise only the affected portion of its reasoning while preserving deductions that remain valid.
    \item \textbf{Necessity judgment.} The model is first given a complete set of information sufficient to derive the correct answer. It is then asked to identify pieces of information that it believes are redundant, that is, observations that can be removed without changing the final conclusion. The goal is not merely to solve the original problem, but to infer a minimal supporting set for the answer according to the model's own dependency analysis. 
\end{itemize}
For counterfactual revision, we measure the success-rate drop relative to the clean backbone. For the necessity-judgment probe, we measure how many queried premises the model judges removable, and whether the pruned evidence set still preserves the correct answer.

\subsection{Benchmark Instantiation and Construction Pipeline}
\label{sec:instantiation}

We instantiate the framework as a benchmark suite of 474 multi-turn interactive games spanning four canonical data structures and three inference modes, with five fixed configuration search spaces corresponding to five standardized difficulty levels. The difficulty is parameterized primarily by monotonically increasing the size of the latent search space (e.g., cardinality of sets, length of sequences, depth of trees, and number of nodes/edges in graphs). Table~\ref{tab:game_counts} summarizes the distribution of games across categories.

Constructing a benchmark of this scale presents a methodological tension. Purely manual construction provides high fidelity but is prohibitively labor-intensive, while unconstrained LLM generation often produces repetitive, inconsistent, or poorly specified games. We therefore adopt a hybrid human-LLM construction strategy that seeks an effective balance between scalability and control.
\begin{itemize}[leftmargin=*]
    \item \textbf{Rule generation under fine-grained constraints.} We begin by generating candidate game specifications at the level of rules. We define a structured design space by crossing four data-structure families (set, sequence, tree, graph) with three inference modes (deductive, inductive, abductive). Within each cell, candidate games are generated subject to constraints on hidden state, query types, observability, answer type, bounded-budget solvability, and difficulty variation. We also manually enumerate possible reasoning targets and probing styles for each structure. Conditioned on these constraints, multiple frontier models, including GPT-5, Claude 4.5, and Gemini 3, generated an initial pool of over 2000 candidate games.
    \item \textbf{Program generation with templates.} The next stage converts selected rule specifications into executable environments. Here, directly asking an LLM to generate each environment from scratch proved error-prone. To reduce implementation failures, we instead use a templated environment interface and abstract away common functionality that is shared across games, including action parsing, format validation, turn-budget management, response serialization, and answer checking. LLMs then implement only the game-specific logic, including hidden-state initialization, query semantics, and observation generation. We further apply iterative execution-based repair using runtime diagnostics and failed test cases.
    \item \textbf{Difficulty and discrimination filtering.} Using preliminary evaluation with GPT-5, Claude 4.5, and Gemini 3, we remove games that are too easy (success rate above 90\%) or too hard (below 10\%), as well as games with suspicious cross-model discrepancies. After this stage, roughly 550 games remained.
    \item \textbf{Fine-grained validation and manual review.} We then conduct a final pass focused on well-posedness, solvability, rule clarity, and overall benchmark quality. Automated assessment is complemented by manual inspection, through which we removed approximately 70 additional games judged to be ambiguous, ill-calibrated, or otherwise unsuitable for reliable evaluation. The resulting benchmark contains 474 games.
\end{itemize}

\textbf{Construction of robustness and adaptation variants.}
For contextual robustness, we generate three perturbation types. For isomorphic perturbation, LLMs rewrite symbolic entities, relations, and rule descriptions into semantically grounded variants across five domains: education, healthcare, transportation, manufacturing, and law. For intra-game noise injection, we insert irrelevant natural-language sentences into rules and meaningless character-level or semantically vacuous fragments into environment responses while preserving evidential content. For inter-game boundary control, each game is initialized with the full interaction history of another randomly sampled game.

For metacognitive adaptation, we similarly derive two controlled variants. In counterfactual revision, the environment intentionally answers the model's second query incorrectly and explicitly corrects it on the following turn \footnote{This ensures that any subsequent failures are caused by an inability to update beliefs properly rather than by an exhausted budget or an irreversibly broken trajectory.}. In the necessity-judgment probe, we enumerate all valid queries for a game instance, collect the complete evidence pool, ask the model to solve the game from this full information state, and then ask it to identify unnecessary observations. Those observations are removed, and the model is re-evaluated on the pruned evidence set. Because the benchmark consists of executable environments under a unified protocol, all these variants can be generated simply at scale.

\begin{table}[!t]
\centering
\scriptsize
\caption{Number of instantiated games by data structure and reasoning type.}
\label{tab:game_counts}
\begin{tabular}{lcccc}
\toprule
& \textbf{Set} & \textbf{Sequence} & \textbf{Tree} & \textbf{Graph} \\
\midrule
\textbf{Inductive} & 24 & 61 & 68 & 57 \\
\textbf{Abductive} & 24 & 39 & 37 & 47 \\
\textbf{Deductive} & 18 & 36 & 27 & 36 \\
\bottomrule
\end{tabular}
\end{table}

\section{Evaluation}

\subsection{Setup}

\textbf{Models.}
We report results for Qwen3-max, Deepseek-3.2, Claude-4.5, GPT-5.4, Gemini-2.5-flash, Gemini-3.1-flash-lite, and Gemini-3.1-pro on the clean backbone benchmark. For several higher-order stress tests, we focus on a representative subset of models (Qwen3-max, Deepseek-3.2, GPT-5.4, and Gemini-3.1-pro) to control evaluation cost while retaining diversity in capability levels and interaction styles.

\textbf{Metrics.}
We report Success Rate (\%) and Avg.\ Turns, where Avg.\ Turns is computed over successful episodes only. We also report Efficiency = Success Rate / Avg.\ Turns. For robustness and adaptation settings, we report the corresponding Success Rate and its change relative to the clean backbone.
\textit{\textbf{We will open-source all the code upon the acceptance of this paper.}}

% \textbf{Implementation details.}
% All models are evaluated in the same executable environment under the same interaction protocol. When temperature control is available, we set temperature to 0.6.

\subsection{Results and Analysis}

\begin{table}[]
\scriptsize
\centering
\caption{Overall performance on the clean interactive reasoning backbone. Success Rate measures the fraction of successfully solved instances, Avg.\ Turns is computed over successful episodes, and Efficiency is defined as Success Rate / Avg.\ Turns.}
\label{tab:overall_backbone}
\begin{tabular}{@{}cccc@{}}
\toprule
Model                 & Success Rate & Avg.\ Turns & Efficiency \\ \midrule
Qwen3-max             & 43.67    & 10.09      & 4.33       \\
Deepseek-3.2          & 45.49    & 13.63      & 3.34       \\
Claude-4.5            & 73.63    & 12.93      & 5.69       \\
GPT-5.4               & 49.17    & 7.92       & 6.21       \\
Gemini-2.5-flash      & 65.48    & 6.91       & 9.48       \\
Gemini-3.1-flash-lite & 44.08    & 7.36       & 5.99       \\
Gemini-3.1-pro        & 74.21    & 6.77       & 10.96      \\ \bottomrule
\end{tabular}
\end{table}

\begin{table*}[]
\scriptsize
\centering
\caption{Success Rate under clean and contextualized (isomorphic perturbation) conditions across the five fixed difficulty levels. \textit{Drop} denotes the performance change from the clean setting to the contextualized setting, so larger positive values indicate weaker contextual robustness.}
\label{tab:context_by_difficulty}
\begin{tabular}{@{}cccccccc@{}}
\toprule
Difficulty               & Condition    & Qwen3-max & Deepseek-3.2 & GPT-5.4 & Claude-4.5 & Gemini-2.5-flash & Gemini-3.1-pro \\ \midrule
\multirow{3}{*}{1} & Clean  & 58.65     & 59.92        & 60.61   & 81.86      & 74.68            & 81.31          \\
                   & Contextualized & 57.09     & 55.31        & 61.87   & 80.59      & 67.43            & 78.27          \\
                   & Drop    & 1.56      & 4.61         & -1.27   & 1.27       & 7.26             & 3.04           \\ \midrule
\multirow{3}{*}{2} & Clean  & 47.47     & 45.38        & 54.87   & 76.16      & 67.51            & 76.46          \\
                   & Contextualized & 45.91     & 47.40        & 49.04   & 75.32      & 61.14            & 71.10          \\
                   & Drop    & 1.56      & -2.02        & 5.82    & 0.84       & 6.37             & 5.36           \\ \midrule
\multirow{3}{*}{3} & Clean  & 42.41     & 45.40        & 45.43   & 73.21      & 63.50            & 72.53          \\
                   & Contextualized & 40.89     & 42.83        & 44.15   & 69.41      & 57.17            & 67.30          \\
                   & Drop    & 1.52      & 2.57         & 1.28    & 3.80       & 6.33             & 5.23           \\ \midrule
\multirow{3}{*}{4} & Clean  & 37.55     & 40.20        & 44.37   & 69.83      & 62.45            & 72.83          \\
                   & Contextualized & 38.82     & 38.42        & 41.44   & 66.67      & 54.30            & 67.51          \\
                   & Drop    & -1.27     & 1.79         & 2.94    & 3.16       & 8.14             & 5.32           \\ \midrule
\multirow{3}{*}{5} & Clean  & 32.28     & 36.53        & 40.53   & 67.09      & 59.28            & 67.93          \\
                   & Contextualized & 33.76     & 36.16        & 38.53   & 64.35      & 52.83            & 61.39          \\
                   & Drop    & -1.48     & 0.37         & 2.00    & 2.74       & 6.46             & 6.54           \\ \bottomrule
\end{tabular}
\end{table*}

\begin{table*}[htbp]
\scriptsize
\centering
\caption{Fine-grained success rate (\%) of four models across data structures and inference modes on the clean backbone benchmark.}
\label{tab:all_models_breakdown}
\begin{tabular}{@{}cccccccccccccccc@{}}
\toprule
\multicolumn{4}{c}{\textbf{Qwen3-max}} & \multicolumn{4}{c}{\textbf{Deepseek-3.2}} & \multicolumn{4}{c}{\textbf{GPT-5.4}} & \multicolumn{4}{c}{\textbf{Gemini-3.1-pro}} \\
\cmidrule(lr){1-4} \cmidrule(lr){5-8} \cmidrule(lr){9-12} \cmidrule(lr){13-16}
\textbf{DS} & Ind. & Ab. & Ded. & \textbf{DS} & Ind. & Ab. & Ded. & \textbf{DS} & Ind. & Ab. & Ded. & \textbf{DS} & Ind. & Ab. & Ded. \\ \midrule
Set       & 44.17 & 26.67 & 40.00 & Set       & 44.26 & 25.00 & 37.11 & Set       & 49.57 & 24.14 & 44.57 & Set       & 62.50 & 57.50 & 71.58 \\
Seq.  & 30.49 & 34.87 & 68.92 & Seq.  & 35.18 & 35.23 & 67.11 & Seq.  & 40.96 & 32.46 & 80.90 & Seq.  & 70.16 & 63.08 & 86.76 \\
Tree      & 42.73 & 35.14 & 65.38 & Tree      & 46.88 & 33.16 & 63.16 & Tree      & 54.66 & 38.33 & 63.49 & Tree      & 69.09 & 72.43 & 75.38 \\
Graph     & 50.88 & 38.72 & 71.14 & Graph     & 63.45 & 35.74 & 81.40 & Graph     & 55.11 & 47.75 & 64.62 & Graph     & 64.21 & 74.89 & 85.43 \\
\bottomrule
\end{tabular}
\end{table*}

\begin{table}[] % 话题转换
\scriptsize
\centering
\caption{Success Rate before and after inter-game boundary control, where a new game is introduced without resetting the prior interaction context.}
\label{tab:boundary_control}
\begin{tabular}{@{}ccc@{}}
\toprule
Model          & Before & After \\ \midrule
Qwen3-max      & 43.67  & 44.39 \\
Deepseek-3.2   & 45.49  & 45.84 \\
GPT-5.4        & 49.17  & 49.66 \\
Gemini-3.1-pro & 74.21  & 77.47 \\ \bottomrule
\end{tabular}
\end{table}

\begin{table}[] % 噪声
\scriptsize
\centering
\caption{Success Rate before and after intra-game noise injection, where irrelevant or non-informative content is inserted into rules or environment responses.}
\label{tab:noise_injection}
\begin{tabular}{@{}ccc@{}}
\toprule
Model          & Before & After \\ \midrule
Qwen3-max      & 43.67  & 40.34 \\
Deepseek-3.2   & 45.49  & 44.19 \\
GPT-5.4        & 49.17  & 42.84 \\
Gemini-3.1-pro & 74.21  & 70.30 \\ \bottomrule
\end{tabular}
\end{table}

\begin{table}[] % 纠正
\scriptsize
\centering
\caption{Success Rate before and after counterfactual revision, where one earlier observation is revised during the interaction.}
\label{tab:counterfactual_correction}
\begin{tabular}{@{}ccc@{}}
\toprule
Model          & Before & After \\ \midrule
Qwen3-max      & 43.67  & 35.99 \\
Deepseek-3.2   & 45.49  & 40.21 \\
GPT-5.4        & 49.17  & 36.69 \\
Gemini-3.1-pro & 74.21  & 57.56 \\ \bottomrule
\end{tabular}
\end{table}

\begin{table*}[]
\scriptsize
\centering
\caption{Results on the necessity-judgment probe. \textit{Clean Backbone} is the success rate on the clean interactive backbone. \textit{Complete Evidence} is the success rate when the model is given the full evidence pool. \textit{After Pruning} is the success rate after removing observations that the model judged unnecessary. \textit{Pruning Ratio} is the average fraction of observations removed by the model.}
\label{tab:sufficiency_necessity}
\begin{tabular}{@{}ccccc@{}}
\toprule
Model & Clean Backbone & Complete Evidence & After Pruning & Pruning Ratio \\ \midrule
Qwen3-max      & 43.67  & 46.25                        & 32.97                          & 0.62                   \\
Deepseek-3.2   & 45.49  & 42.48                        & 18.21                          & 0.65                   \\
GPT-5.4        & 49.17  & 54.80                        & 37.89                          & 0.76                   \\
Gemini-3.1-pro & 74.21  & 70.76                        & 48.87                          & 0.83                   \\ \bottomrule
\end{tabular}
\end{table*}

\paragraph{Overall performance on the clean backbone.}
Table~\ref{tab:overall_backbone} reports average results on the clean interactive reasoning backbone across the five fixed difficulty levels. The benchmark is strongly discriminative: model accuracy ranges from 43.67\% to 74.21\%. Gemini-3.1-pro achieves the best overall performance, with 74.21\% accuracy, the lowest average turn count among the strongest models (6.77), and the highest efficiency score (10.96). Claude-4.5 reaches a comparable accuracy of 73.63\%, but requires nearly twice as many turns (12.93), yielding a much lower efficiency score of 5.69. This contrast suggests that both models can solve a large fraction of games, but Gemini-3.1-pro does so more directly and with stronger interaction efficiency. GPT-5.4 occupies a different regime where the accuracy is only moderate (49.17\%), but it is relatively concise in interaction (7.92 turns), indicating a more aggressive strategy that is efficient when correct but less reliable overall. Qwen3-max, Deepseek-3.2, and Gemini-3.1-flash-lite cluster in a lower-performance band around 44--45\% accuracy. Moreover, we see that Deepseek-3.2 uses substantially more turns than the others, suggesting that additional interaction alone is insufficient to close the reasoning gap.

\paragraph{Difficulty scaling and contextual robustness.}
Table~\ref{tab:context_by_difficulty} compares clean and contextualized performance across five difficulty levels. As expected, success rates decrease as difficulty increases, confirming that the fixed configuration spaces provide a meaningful difficulty gradient. For example, Gemini-3.1-pro drops from 81.31\% at difficulty 1 to 67.93\% at difficulty 5 in the clean setting, while Claude-4.5 shows a similar decline from 81.86\% to 67.09\%.

Isomorphic contextualization introduces an additional robustness challenge. Although the underlying latent problem is unchanged, most models suffer performance drops under semantically enriched surface forms. Claude-4.5 is the most stable model in this setting, with small drops across all difficulty levels. Gemini-2.5-flash and Gemini-3.1-pro, despite strong clean performance, are more sensitive to contextualization, often losing more than five percentage points. Some models occasionally improve under contextualization, suggesting that richer semantic descriptions may provide useful priors in isolated cases. However, the dominant trend is degradation, indicating that preserving abstract reasoning under surface-level semantic shifts remains nontrivial.

\paragraph{Structure- and inference-level breakdown.}
Table~\ref{tab:all_models_breakdown} shows clear differences across inference modes and data structures. Deductive reasoning is consistently the strongest setting for all models. We attribute this to the dominant reasoning-oriented training data for current LLMs, such as mathematics and code, which mainly emphasize deriving conclusions from explicit premises. In contrast, inductive and abductive reasoning require pattern formation or explanation search from partial evidence, and are less directly supported by such training signals.

We also observe that set-based tasks are often harder than tree and graph tasks. Unlike trees and graphs, which provide explicit structural relations, sets offer fewer organizational cues and require models to maintain more unordered information internally. This may make evidence tracking and hypothesis management more difficult.

\paragraph{Boundary control and noise robustness.}
Tables~\ref{tab:boundary_control} and Table~\ref{tab:noise_injection} examine two additional contextual robustness probes. Under inter-game boundary control, where a new game begins without clearing the previous interaction history, performance remains stable and even improves slightly for all tested models. This suggests that stale context alone is not a major obstacle, i.e., models can usually recognize the new task boundary and reinitialize their reasoning state. In contrast, intra-game noise injection consistently reduces performance. GPT-5.4 drops from 49.17\% to 42.84\%, and Gemini-3.1-pro drops from 74.21\% to 70.30\%. Qwen3-max and Deepseek-3.2 also decline. The contrast between boundary control and noise injection indicates that the main robustness challenge is not simply long-context interference, but the ability to filter irrelevant content while tracking active evidence within the current game.

\paragraph{Counterfactual revision.}
Table~\ref{tab:counterfactual_correction} shows that counterfactual revision is substantially more difficult than the contextual perturbations. Revising one earlier observation causes large drops for all models: Qwen3-max decreases from 43.67\% to 35.99\%, Deepseek-3.2 from 45.49\% to 40.21\%, etc. These results suggest that current models often fail to update their internal reasoning state after evidence changes. Belief revision under partial observability therefore appears to be a qualitatively harder capability than robustness to surface-level contextual variation.

\paragraph{Necessity judgment.}
Table~\ref{tab:sufficiency_necessity} evaluates whether models can distinguish necessary evidence from merely available evidence. Providing the complete evidence pool does not uniformly improve performance: GPT-5.4 and Qwen3-max benefit modestly, whereas Deepseek-3.2 and Gemini-3.1-pro perform slightly worse than in the clean interactive setting. This indicates that full information does not remove the need for coherent evidence integration. After pruning observations that the model itself judged unnecessary, success rates drop sharply for all models. Gemini-3.1-pro falls from 70.76\% to 48.87\%, GPT-5.4 from 54.80\% to 37.89\%, Deepseek-3.2 from 42.48\% to 18.21\%, and Qwen3-max from 46.25\% to 32.97\%. At the same time, pruning ratios are high, ranging from 0.62 to 0.83. Thus, models frequently remove a large fraction of observations while mistakenly discarding information required for the conclusion. This gap suggests that current LLMs are better at using evidence when it is present than at identifying which evidence is indispensable.

\begin{table}[t]
\scriptsize
\centering
\caption{Failure attribution. \textit{Timeout} denotes episodes that exceed the maximum turn budget, and \textit{FormatError} denotes episodes in which the model fails to follow the required query or submission format.}
\label{tab:failure_modes}
\begin{tabular}{@{}ccc@{}}
\toprule
Model & Timeout & FormatError \\ \midrule
Qwen3-max      & 15 & 171 \\
Deepseek-3.2   & 20 & 195 \\
GPT-5.4        & 2  & 122 \\
Gemini-3.1-pro & 1  & 316 \\ \bottomrule
\end{tabular}
\end{table}

\paragraph{Failure modes.}
Table~\ref{tab:failure_modes} shows that timeouts are rare across models, suggesting that the turn budget is generally sufficient. The dominant non-solution failure mode is instead \textit{FormatError}. This highlights a distinctive challenge of interactive evaluation: models must not only reason over hidden states, but also maintain reliable protocol compliance across multiple turns. The effect is especially visible for Gemini-3.1-pro, which has very few timeouts but many format-related failures.

\section{Conclusion}

We introduced a hierarchical framework and benchmark for interactive reasoning that evaluates LLMs as active agents under partial observability.
The benchmark combines a clean interactive backbone with controlled probes for contextual robustness and metacognitive adaptation.
Our results show that current frontier models already exhibit nontrivial interactive reasoning ability, but their performance remains uneven across settings.
More broadly, we hope this benchmark helps future work on reasoning diagnostics, training-time intervention, tool-augmented agents, and interactive reinforcement learning, because the environment is executable and the interaction protocol is explicit.

\section*{Limitations}

Several limitations of this work should be acknowledged. First, the five fixed configuration search spaces, though carefully calibrated for difficulty scaling, cannot fully capture the continuous and unbounded nature of reasoning challenges encountered in practice. Second, the interactive protocol employs a structured XML-based action format that, while ensuring reliable parsing and evaluation consistency, may oversimplify the interaction dynamics compared to free-form natural language communication, potentially overestimating models' ability to maintain protocol compliance in less constrained settings.

\section*{Ethics Statement}

This work introduces a benchmark for evaluating interactive reasoning in LLMs and does not involve any human subjects, user studies, or collection of personally identifiable information. All benchmark games are constructed from abstract symbolic primitives (sets, sequences, trees, and graphs) across deductive, inductive, and abductive inference modes, and do not contain any offensive, biased, or culturally sensitive content. The isomorphic perturbation variants are drawn from five neutral domains (education, healthcare, transportation, manufacturing, and law) and are designed solely to test structural reasoning invariance rather than to make any substantive claims about these domains.

% \section*{Acknowledgments}

% This document has been adapted
% by Steven Bethard, Ryan Cotterell and Rui Yan
% from the instructions for earlier ACL and NAACL proceedings, including those for
% ACL 2019 by Douwe Kiela and Ivan Vuli\'{c},
% NAACL 2019 by Stephanie Lukin and Alla Roskovskaya,
% ACL 2018 by Shay Cohen, Kevin Gimpel, and Wei Lu,
% NAACL 2018 by Margaret Mitchell and Stephanie Lukin,
% Bib\TeX{} suggestions for (NA)ACL 2017/2018 from Jason Eisner,
% ACL 2017 by Dan Gildea and Min-Yen Kan,
% NAACL 2017 by Margaret Mitchell,
% ACL 2012 by Maggie Li and Michael White,
% ACL 2010 by Jing-Shin Chang and Philipp Koehn,
% ACL 2008 by Johanna D. Moore, Simone Teufel, James Allan, and Sadaoki Furui,
% ACL 2005 by Hwee Tou Ng and Kemal Oflazer,
% ACL 2002 by Eugene Charniak and Dekang Lin,
% and earlier ACL and EACL formats written by several people, including
% John Chen, Henry S. Thompson and Donald Walker.
% Additional elements were taken from the formatting instructions of the \emph{International Joint Conference on Artificial Intelligence} and the \emph{Conference on Computer Vision and Pattern Recognition}.

\bibliography{custom}

@article{ref5,
  author       = {Karl Cobbe and
                  Vineet Kosaraju and
                  Mohammad Bavarian and
                  Mark Chen and
                  Heewoo Jun and
                  Lukasz Kaiser and
                  Matthias Plappert and
                  Jerry Tworek and
                  Jacob Hilton and
                  Reiichiro Nakano and
                  Christopher Hesse and
                  John Schulman},
  title        = {Training Verifiers to Solve Math Word Problems},
  journal      = {CoRR},
  volume       = {abs/2110.14168},
  year         = {2021},
  url          = {https://arxiv.org/abs/2110.14168},
  eprinttype   = {arXiv},
  eprint       = {2110.14168},
  bibsource    = {dblp computer science bibliography, https://dblp.org}
}

@inproceedings{ref9,
  author       = {Weihao Yu and
                  Zihang Jiang and
                  Yanfei Dong and
                  Jiashi Feng},
  title        = {ReClor: {A} Reading Comprehension Dataset Requiring Logical Reasoning},
  booktitle    = {8th International Conference on Learning Representations, {ICLR} 2020,
                  Addis Ababa, Ethiopia, April 26-30, 2020},
  publisher    = {OpenReview.net},
  year         = {2020},
  url          = {https://openreview.net/forum?id=HJgJtT4tvB},
  bibsource    = {dblp computer science bibliography, https://dblp.org}
}

@article{ref10,
  title={Logiqa: A challenge dataset for machine reading comprehension with logical reasoning},
  author={Liu, Jian and Cui, Leyang and Liu, Hanmeng and Huang, Dandan and Wang, Yile and Zhang, Yue},
  journal={arXiv preprint arXiv:2007.08124},
  year={2020}
}

@article{ref18,
  title={Measuring mathematical problem solving with the math dataset},
  author={Hendrycks, Dan and Burns, Collin and Kadavath, Saurav and Arora, Akul and Basart, Steven and Tang, Eric and Song, Dawn and Steinhardt, Jacob},
  journal={arXiv preprint arXiv:2103.03874},
  year={2021}
}

@inproceedings{ref20,
  title={Mathvista: Evaluating mathematical reasoning of foundation models in visual contexts},
  author={Lu, Pan and Bansal, Hritik and Xia, Tony and Liu, Jiacheng and Li, Chunyuan and Hajishirzi, Hannaneh and Cheng, Hao and Chang, Kai-Wei and Galley, Michel and Gao, Jianfeng},
  booktitle={International Conference on Learning Representations},
  volume={2024},
  pages={23439--23554},
  year={2024}
}

@article{ref22,
  title={Humanity's last exam},
  author={Phan, Long and Gatti, Alice and Han, Ziwen and Li, Nathaniel and Hu, Josephina and Zhang, Hugh and Zhang, Chen Bo Calvin and Shaaban, Mohamed and Ling, John and Shi, Sean and others},
  journal={arXiv preprint arXiv:2501.14249},
  year={2025}
}

@article{ref23,
  title={Frontiermath: A benchmark for evaluating advanced mathematical reasoning in ai},
  author={Glazer, Elliot and Erdil, Ege and Besiroglu, Tamay and Chicharro, Diego and Chen, Evan and Gunning, Alex and Olsson, Caroline Falkman and Denain, Jean-Stanislas and Ho, Anson and Santos, Emily de Oliveira and others},
  journal={arXiv preprint arXiv:2411.04872},
  year={2024}
}

@article{ref26,
  title={Evaluating large language models trained on code},
  author={Chen, Mark and Tworek, Jerry and Jun, Heewoo and Yuan, Qiming and Pinto, Henrique Ponde De Oliveira and Kaplan, Jared and Edwards, Harri and Burda, Yuri and Joseph, Nicholas and Brockman, Greg and others},
  journal={arXiv preprint arXiv:2107.03374},
  year={2021}
}

@article{ref27,
  title={Measuring coding challenge competence with apps},
  author={Hendrycks, Dan and Basart, Steven and Kadavath, Saurav and Mazeika, Mantas and Arora, Akul and Guo, Ethan and Burns, Collin and Puranik, Samir and He, Horace and Song, Dawn and others},
  journal={arXiv preprint arXiv:2105.09938},
  year={2021}
}

@inproceedings{ref30,
  title={Swe-bench: Can language models resolve real-world github issues?},
  author={Jimenez, Carlos E and Yang, John and Wettig, Alexander and Yao, Shunyu and Pei, Kexin and Press, Ofir and Narasimhan, Karthik},
  booktitle={International Conference on Learning Representations},
  volume={2024},
  pages={54107--54157},
  year={2024}
}

@article{ref36,
  title={Judging llm-as-a-judge with mt-bench and chatbot arena},
  author={Zheng, Lianmin and Chiang, Wei-Lin and Sheng, Ying and Zhuang, Siyuan and Wu, Zhanghao and Zhuang, Yonghao and Lin, Zi and Li, Zhuohan and Li, Dacheng and Xing, Eric and others},
  journal={Advances in neural information processing systems},
  volume={36},
  pages={46595--46623},
  year={2023}
}

@inproceedings{ref38,
  title={Mt-bench-101: A fine-grained benchmark for evaluating large language models in multi-turn dialogues},
  author={Bai, Ge and Liu, Jie and Bu, Xingyuan and He, Yancheng and Liu, Jiaheng and Zhou, Zhanhui and Lin, Zhuoran and Su, Wenbo and Ge, Tiezheng and Zheng, Bo and others},
  booktitle={Proceedings of the 62nd Annual Meeting of the Association for Computational Linguistics (Volume 1: Long Papers)},
  pages={7421--7454},
  year={2024}
}

@inproceedings{ref40,
  title={Multichallenge: A realistic multi-turn conversation evaluation benchmark challenging to frontier llms},
  author={Deshpande, Kaustubh and Sirdeshmukh, Ved and Mols, Johannes Baptist and Jin, Lifeng and Hernandez-Cardona, Ed-Yeremai and Lee, Dean and Kritz, Jeremy and Primack, Willow E and Yue, Summer and Xing, Chen},
  booktitle={Findings of the Association for Computational Linguistics: ACL 2025},
  pages={18632--18702},
  year={2025}
}

@article{ref41,
  title={Mtr-bench: A comprehensive benchmark for multi-turn reasoning evaluation},
  author={Li, Xiaoyuan and Bao, Keqin and Ma, Yubo and Li, Moxin and Wang, Wenjie and Men, Rui and Zhang, Yichang and Feng, Fuli and Liu, Dayiheng and Lin, Junyang},
  journal={arXiv preprint arXiv:2505.17123},
  year={2025}
}

@article{ref42,
  title={TurnBench-MS: A benchmark for evaluating multi-turn, multi-step reasoning in large language models},
  author={Zhang, Yiran and Wang, Mo and Li, Xiaoyang and Ren, Kaixuan and Zhu, Chencheng and Naseem, Usman},
  journal={Findings of the Association for Computational Linguistics: EMNLP},
  pages={19892--19924},
  year={2025}
}

@article{Multi-Turn-Puzzles,
  title={Multi-turn puzzles: Evaluating interactive reasoning and strategic dialogue in llms},
  author={Badola, Kartikeya and Simon, Jonathan and Hosseini, Arian and Carthy, Sara Marie Mc and Munkhdalai, Tsendsuren and Goyal, Abhimanyu and Ko{\v{c}}isk{\`y}, Tom{\'a}{\v{s}} and Upadhyay, Shyam and Fatemi, Bahare and Kazemi, Mehran},
  journal={arXiv preprint arXiv:2508.10142},
  year={2025}
}

@article{GTBENCH,
  title={Gtbench: Uncovering the strategic reasoning capabilities of llms via game-theoretic evaluations},
  author={Duan, Jinhao and Zhang, Renming and Diffenderfer, James and Kailkhura, Bhavya and Sun, Lichao and Stengel-Eskin, Elias and Bansal, Mohit and Chen, Tianlong and Xu, Kaidi},
  journal={Advances in Neural Information Processing Systems},
  volume={37},
  pages={28219--28253},
  year={2024}
}

@inproceedings{GameArena,
  title={Gamearena: Evaluating llm reasoning through live computer games},
  author={Hu, Lanxiang and Li, Qiyu and Xie, Anze and Jiang, Nan and Stoica, Ion and Jin, Haojian and Zhang, Hao},
  booktitle={International Conference on Learning Representations},
  volume={2025},
  pages={33278--33309},
  year={2025}
}

@article{deepseekv3,
  title={Deepseek-v3 technical report},
  author={Liu, Aixin and Feng, Bei and Xue, Bing and Wang, Bingxuan and Wu, Bochao and Lu, Chengda and Zhao, Chenggang and Deng, Chengqi and Zhang, Chenyu and Ruan, Chong and others},
  journal={arXiv preprint arXiv:2412.19437},
  year={2024}
}

@article{gemini,
  title={Gemini 2.5: Pushing the frontier with advanced reasoning, multimodality, long context, and next generation agentic capabilities},
  author={Comanici, Gheorghe and Bieber, Eric and Schaekermann, Mike and Pasupat, Ice and Sachdeva, Noveen and Dhillon, Inderjit and Blistein, Marcel and Ram, Ori and Zhang, Dan and Rosen, Evan and others},
  journal={arXiv preprint arXiv:2507.06261},
  year={2025}
}

@inproceedings{bang2025hallulens,
  title={Hallulens: Llm hallucination benchmark},
  author={Bang, Yejin and Ji, Ziwei and Schelten, Alan and Hartshorn, Anthony and Fowler, Tara and Zhang, Cheng and Cancedda, Nicola and Fung, Pascale},
  booktitle={Proceedings of the 63rd Annual Meeting of the Association for Computational Linguistics (Volume 1: Long Papers)},
  pages={24128--24156},
  year={2025}
}

@article{yin2025reasoning,
  title={The reasoning trap: How enhancing LLM reasoning amplifies tool hallucination},
  author={Yin, Chenlong and Sha, Zeyang and Cui, Shiwen and Meng, Changhua and Li, Zechao},
  journal={arXiv preprint arXiv:2510.22977},
  year={2025}
}
% \bibliographystyle{acl}
% \bibliography{custom}

\appendix

\section{Structural Coverage and Inference Type Definitions}

The framework covers four canonical data structures, chosen for their foundational role in symbolic reasoning:
\begin{itemize} 
	\item \textbf{Set:} Unordered collection of distinct elements. Reasoning tasks involve membership, subset relations, intersections, and cardinality constraints.
	\item \textbf{Sequence:} Ordered list of elements. Reasoning tasks involve ordering, adjacency, indexing, pattern completion, and transformation rules.
	\item \textbf{Tree:} Hierarchical structure with parent-child relationships. Reasoning tasks involve ancestry, subtree properties, path constraints, and recursive definitions.
	\item \textbf{Graph:} Network of nodes connected by edges. Reasoning tasks involve connectivity, paths, cycles, reachability, and constraint propagation.
\end{itemize}

Each data structure supports tasks across three inference types, defined as follows:
\begin{itemize}
	\item \textbf{Deductive inference:} The model is provided with complete rules and initial conditions. The hidden state is fully determined by these premises, and the task is to derive the necessary conclusion. Success requires correct rule application rather than hypothesis generation.
	\item \textbf{Inductive inference:} The model is provided with partial observations. The hidden state is a latent rule or pattern that must be inferred from examples. Multiple candidate rules may be consistent with the observed data, and the model must identify the one intended by the game designer.
	\item \textbf{Abductive inference:} The model is provided with an observed outcome. The hidden state is the most plausible cause or configuration that could have produced this outcome. Multiple hidden states may be consistent with the same observation, and the model must reason backward to identify the most likely explanation.
\end{itemize}

\section{Implementation Overview}
\label{app:implementation}

All benchmark instances are implemented under a shared executable \texttt{Game} interface. The common runtime handles prompt construction, XML-based action parsing, transcript management, and episode status updates. Each concrete game only needs to define three components: (1) how the hidden state is initialized, (2) how the environment answers valid queries, and (3) how final answers are verified.

\textbf{Shared runtime.}
At initialization, a game instance is created from a configuration specifying one of five fixed difficulty levels and the context condition. The environment first constructs the hidden state, then selects either the clean rule template or one of its contextualized rewrites, and finally appends the rendered rule prompt to the interaction history. During interaction, each model response is parsed into either a query action or a final answer. Query actions are executed by the environment, while final answers are passed to a task-specific verifier.

\begin{lstlisting}[language=Python,basicstyle=\ttfamily\small]
class Game(ABC):

    def initialize_game(self):

    def step(self, response):
    # logic flow

    def evaluate(self, parsed_info):
    # verify final answer

    def produce(self, parsed_info):
    # make a response to a query

    def get_all_possible_queries(self): 
    # enumerate legal queries
    # for pruning-based probes

    def cf_make_wrong(self, correct):
    # generate a controlled
    # wrong observation
\end{lstlisting}

The same abstraction also supports the two higher-order probes. \emph{Counterfactual revision} is implemented by wrapping the ordinary response function: the environment returns one controlled wrong observation and then explicitly corrects it on the next turn. \emph{Necessity-judgment probing} is implemented by exposing a function that enumerates all legal queries and their answers for a fixed hidden instance, which allows us to construct a complete evidence pool and then remove the observations that the model judges unnecessary. In other words, both probes are realized through additional functions on top of the same executable game class, rather than through separate benchmark implementations.

\textbf{Rule Example.}
A minimal example is a hidden-number game. The environment secretly chooses one number from $\{1,2,3,4\}$. The model may ask whether the number is odd, whether it is greater than 2, or whether it equals a specific candidate. It must then submit the hidden number.
A shortened rule prompt is as follows:
\begin{quote}\small
I have selected a hidden number from \{1,2,3,4\}. Your goal is to find the number.
You may ask one question per turn using the following format:
\texttt{<query\_odd></query\_odd>} \\
\texttt{<query\_greater>2</query\_greater>} \\
\texttt{<query\_equal>3</query\_equal>}
When you know the answer, submit:
\texttt{<answer>3</answer>}
\end{quote}
This design keeps the benchmark modular: all tasks share the same runtime protocol, while individual games only define hidden-state construction, query semantics, and answer verification. It also makes contextualization, \emph{counterfactual revision}, and \emph{necessity-judgment} pruning easy to implement as lightweight wrappers over the same executable environment.

\end{document}